# Interpretable Machine Learning for Antepartum Prediction of Pregnancy-Associated Thrombotic Microangiopathy Using Routine Longitudinal Laboratory Data

**Authors**

Chuanchuan Sun[1,2,3]†, Zhen Yu[3]†, Qin Fan[2,4], Qingchao Chen[3,1]*, Feng Yu[1,2,3]*

**Affiliations**

[1]Public Health and Preventive Medicine of Peking University School of Public Health, Peking University, Beijing, 100191, China

[2]Department of Nephrology, Peking University International Hospital, Beijing, 102206, China.

[3]National Institute of Health Data Science, Peking University, Beijing, 100191, China.

[4]Peking University Health Science Center, Peking University, Beijing, 100191, China

Address correspondence to: yufengevert1@sina.com; qingchao.chen@pku.edu.cn.

†These authors contributed equally to this work.

**Abstract**

**Background:** Pregnancy-associated thrombotic microangiopathy (P-TMA) is rare but life-threatening. Early risk prediction before overt clinical presentation remains challenging, as the associated laboratory abnormalities are subtle, multidimensional, and frequently masked by common physiological changes such as gestational thrombocytopenia and pregnancy-related proteinuria, thus overlapping heavily with benign obstetric and renal conditions. This complexity is poorly captured by univariate or rule-based approaches; however, it is addressable by machine learning, which can extract latent, time-dependent risk signatures from longitudinal clinical tests. **Methods:** This retrospective study included 300 pregnancies comprising 142 P-TMA cases and 158 controls. After exclusion of identifiers and non-informative variables, 146 longitudinal laboratory predictors were retained. Participants were divided into a training cohort (80%) and a held-out test cohort (20%) using stratified sampling. Five algorithms were evaluated: logistic regression, support vector machine with radial basis function kernel, random forest, extra trees, and gradient boosting. The final model was selected by mean cross-validated AUROC, refitted on the full training cohort, and evaluated once in the held-out test cohort. Interpretability analyses examined global feature importance and distributional patterns of leading predictors. **Results:** Gradient boosting was prespecified by cross-validation in the training cohort. The model achieved an AUROC of 0.872 (95% CI: 0.769–0.952) and an AUPRC of 0.883 (95% CI: 0.780–0.959) in a held-out test cohort, with sensitivity of 0.750 and specificity of 0.812. Key predictors included first-trimester renal markers (cystatin C and uric acid at week 6), second-trimester inflammatory shifts (neutrophils at week 16 and neutrophil-to-lymphocyte ratio at week 24), gestational biochemical injury markers (lactate dehydrogenase), and third-trimester renal impairment with anemia (creatinine at week 24, hemoglobin at week 32). **Conclusions:** Longitudinal clinical laboratory tests obtained during routine care contained informative and clinically plausible signals for P-TMA risk. Notably, cystatin C at week 6 showed promise as an early monitoring indicator.

## 1 Introduction

Pregnancy-associated thrombotic microangiopathy (P-TMA) encompasses a group of rare yet catastrophic perinatal syndromes, most notably thrombotic thrombocytopenic purpura (TTP) and complement-mediated hemolytic uremic syndrome (aHUS), which typically manifest in the late third trimester and postpartum period (predominantly between 34 weeks of gestation and 3 months postpartum)[1,2]. These disorders share a triad of pathological features—microangiopathic hemolytic anemia, consumptive thrombocytopenia, and ischemic organ damage caused by microvascular thrombosis—with renal involvement being particularly prominent[1-4]. With an estimated incidence of only 1 per 200,000 pregnancies, P-TMA is nevertheless rapidly progressive, often culminating in deteriorating renal function, multiorgan failure, fetal loss, or maternal death[5-7]. By the time the classic "pentad" or overt hemolysis, proteinuria, and renal failure emerge, the therapeutic window has typically narrowed dramatically. Hence, early warning and risk prediction before clinical deterioration represent critical leverage points for improving maternal-fetal outcomes[1,2].

Despite its urgency, early P-TMA recognition remains a formidable clinical challenge. Prodromal symptoms and laboratory abnormalities are notoriously nonspecific, leading to frequent confusion with other pregnancy-related conditions, including STEC-HUS, Drug-Mediated TMA, acute fatty liver of pregnancy (AFLP), active systemic lupus erythematosus (SLE), and antiphospholipid syndrome (APS)[6,8,9]. These entities show marked overlap in hematological, hepatic, renal, and inflammatory parameters. Definitive tests—ADAMTS13 activity, complement genetic analysis, and anti-factor H antibodies—are not only time-consuming but also unavailable in many primary or emergency settings, precluding their use as universal screening tools[10,11]. Compounding the challenge, normal pregnancy itself induces dynamic physiological changes—including a progressive ~50% decline in ADAMTS13 activity and mild thrombocytopenia—making it exceedingly difficult to distinguish pathological signals from physiological alterations using univariate or rule-based approaches[12,13]. Moreover, given the rarity of these disorders, large-scale randomized trials are infeasible, and high-quality evidence-based diagnostic guidelines are consequently lacking[2]. Machine learning offers a promising solution to these intertwined challenges: it can automatically learn nonlinear interactions among dozens of laboratory parameters, capture time-dependent risk trajectories across gestational weeks, and derive predictive signatures directly from routine clinical data without relying on pre-existing guidelines or expert rules, thereby enabling early risk identification in a manner unattainable by traditional approaches.

The proliferation of electronic health records (EHR) and advances in data science have deepened the application of clinical prediction models in obstetrics. Machine learning algorithms, particularly gradient boosting (e.g., XGBoost, LightGBM) and deep learning models, substantially outperform traditional logistic regression in handling high-dimensional, nonlinear, and time-series laboratory data[14-16]. For instance, in predicting preeclampsia, gestational diabetes, and preterm birth, models integrating longitudinal blood pressure, biochemical markers, and ultrasound parameters have achieved moderate to high discriminative accuracy[17-21]. Crucially, the advent of explainable artificial intelligence (XAI)—exemplified by SHapley Additive exPlanations (SHAP) and feature importance analysis—has made it possible to deconstruct these otherwise opaque models, mapping their decision paths back to specific clinical variables and revealing underlying pathophysiological mechanisms[22-24].

To date, machine learning prediction models covering the full spectrum of P-TMA remain entirely undeveloped. Existing work is confined to a general TMA diagnostic model for hospitalized patients—designed for diagnosis at admission, not for risk screening in asymptomatic pregnant women[25]. Meanwhile, single-subtype studies have focused on conditions such as HELLP syndrome or TTP in selected patient cohorts[26,27], without addressing the complete, heterogeneous P-TMA spectrum. Here, we construct an explainable machine learning framework integrating longitudinal routine laboratory data from gestational weeks 6 to 32 (including multi-time-point blood, biochemical, and urinary indices) to develop and internally validate a gradient boosting model for P-TMA risk prediction. The algorithm captures high-dimensional nonlinear feature interactions and time-dependent risk trajectories—patterns imperceptible to human analysis. This enables effective risk prediction before clinical diagnosis, informing high-risk management and intrapartum monitoring. This study provides the first clinically actionable strategy for early, full-spectrum P-TMA risk during the asymptomatic antepartum period.

## 2 Methods

## 2.1 Study design and participants

We conducted a retrospective clinical prediction modeling study at Peking University International Hospital. Participant selection is summarized in Figure 1. The final analytic cohort included 300 pregnancies, comprising 142 patients with P-TMA and 158 controls. P-TMA status was defined according to compatible thrombotic microangiopathy features during pregnancy or the postpartum period, including thrombocytopenia, microangiopathic hemolysis, and organ involvement, with renal injury considered a key manifestation[2]. The study focused on risk prediction from routinely available longitudinal clinical laboratory tests.

The modeling objective was risk rather than etiological classification. Predictors were therefore restricted to variables obtainable from routine clinical laboratory testing. Patient identifiers and variables that could directly encode outcome status or post-event information were excluded before model development.

## 2.2 Laboratory predictors and preprocessing

Candidate predictors were derived from serial clinical laboratory tests collected during routine care. These included hematologic indices, renal function markers, liver injury markers, urinalysis variables, inflammatory indicators, and related biochemical measurements. Longitudinal timepoint information was retained in the feature names using Roman numerals: VI, XII, XVI, XXIV, XXVIII, XXXand XXXII corresponded to weeks 6, 12, 16, 24, 28, 30 and 32, respectively. For example, CysCVI denotes cystatin C at week 6, UAVI denotes uric acid at week 6, LDHVI denotes lactate dehydrogenase at week 6, and HbXXXII denotes hemoglobin at week 32.

Preprocessing was implemented as a reproducible pipeline. Identifier fields and variables with potential outcome leakage were removed. Object-type columns were parsed to recover numeric clinical values when appropriate, including entries containing units or date annotations. Pure date fields, high-cardinality free-text fields, variables exceeding the prespecified missingness threshold, and constant variables were excluded. After preprocessing, 146 longitudinal laboratory predictors were retained.

All imputation and transformation steps were embedded within the modeling pipeline. Continuous variables were imputed using the median and standardized. Categorical variables, when present, were imputed using the most frequent category and one-hot encoded. These operations were estimated within the training workflow to avoid leakage from the held-out test cohort.

## 2.3 Model development and selection

Participants were divided into a training cohort and a held-out test cohort using an 80:20 stratified split. Five candidate algorithms were evaluated: logistic regression, support vector machine with radial basis function kernel, random forest, extra trees, and gradient boosting. Hyperparameters were optimized using randomized search with five-fold stratified cross-validation within the training cohort.

The final model was selected according to the area under the receiver operating characteristic curve (AUROC) in the training cohort. After selection, the model was refit on the full training cohort and evaluated once in the held-out test cohort. The classification threshold was derived from out-of-fold training predictions using the Youden index. This design preserved the held-out test cohort for final evaluation and avoided post hoc model selection based on test-set performance. Candidate-model test results were reported for transparency, but the test cohort was not used to revise the selected model.

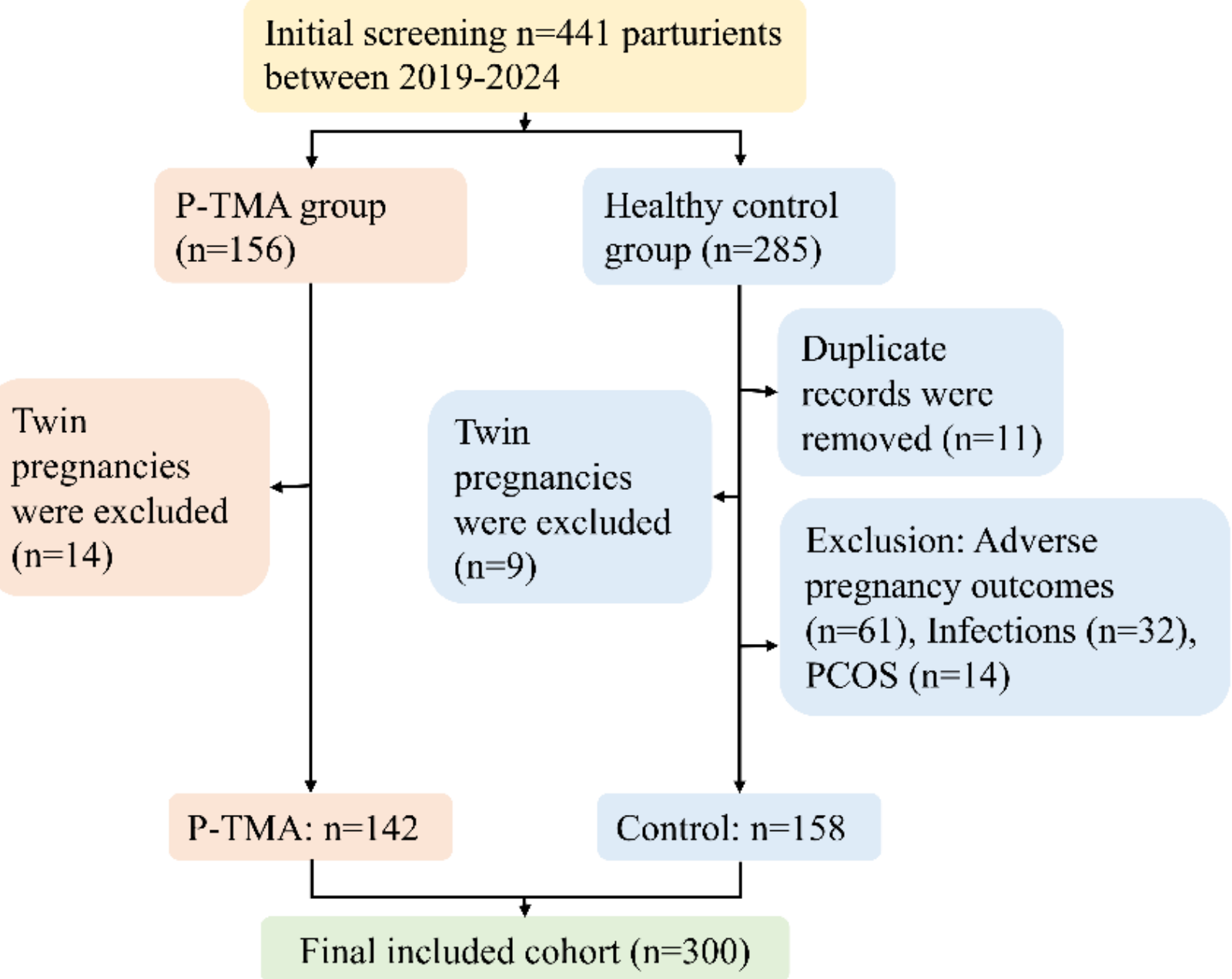

**Figure 1: Flow diagram of study participant selection.** The final analytic cohort included 142 P-TMA cases and 158 controls. P-TMA: pregnancy-associated thrombotic microangiopathy; PCOS: polycystic ovary syndrome.

## 2.4 Performance evaluation

Discrimination was assessed using the AUROC and the area under the precision-recall curve (AUPRC). Threshold-dependent performance was summarized using accuracy, sensitivity, specificity, positive predictive value (PPV), negative predictive value (NPV), and F1 score. Probabilistic accuracy was evaluated using the Brier score, with lower values indicating better agreement between predicted probabilities and observed outcomes. Calibration was assessed visually and using expected calibration error.

Clinical utility was evaluated using decision-curve analysis, which estimates the net benefit of model-guided decision-making across threshold probabilities relative to treat-all and treat-none strategies[28]. Bootstrap resampling with 1000 iterations was used to estimate 95% confidence intervals for held-out test performance metrics.

## 2.5 Model interpretation

Model interpretation focused on whether the selected model relied on clinically plausible laboratory patterns. Global feature importance was used to identify leading predictors. Raw feature codes were retained in the main interpretation figure to preserve the original longitudinal laboratory-feature convention, while clinical aliases were provided in Table 4.

For leading predictors, group-level distributional patterns were further examined. Because some laboratory tests are rounded or have narrow physiologic ranges, simple median differences can underestimate distributional separation. We therefore summarized direction of group difference using signed rank-biserial effect sizes derived from the Mann-Whitney U statistic. Positive values indicated higher distributions in the P-TMA group, whereas negative values indicated lower distributions.

For visualization across laboratory tests with different units and scales, predictor distributions were transformed to robust z-scores based on the median and interquartile range. These interpretation analyses were intended to support clinical plausibility and should not be interpreted as causal evidence.

### 2.6 Statistical analysis and reporting

Continuous variables were summarized as median [interquartile range], and between-group comparisons used the Mann-Whitney U test. Categorical variables were summarized as counts and percentages and compared using Fisher's exact test or the chi-square test as appropriate. Two-sided $P < 0.05$ was considered statistically significant. Analyses were implemented in Python using a reproducible pipeline included in the project archive. Reporting was guided by TRIPOD principles and current recommendations for clinical prediction model development and evaluation[28,29].

## 3 Results

### 3.1 Cohort characteristics and predictors

The final cohort included 300 pregnancies, comprising 142 P-TMA cases and 158 controls (Table 1). The two groups were similar in age and history of adverse pregnancy outcomes. Compared with controls, patients with P-TMA delivered earlier, had higher admission systolic and diastolic blood pressure, higher prepregnancy body mass index, lower newborn birth weight, more frequent adverse fetal outcomes, longer hospital stay, and markedly higher urinary protein and microalbuminuria levels. These differences are consistent with the renal, vascular, and obstetric burden of P-TMA and related hypertensive pregnancy complications.

**Table 1:** Baseline demographic and clinical characteristics of the study cohort.

| Characteristic | Overall (n=300) | Control (n=158) | P-TMA (n=142) | *P* value |
|---|---|---|---|---|
| Age, years | 31 [29, 34] | 31 [29, 34] | 31 [29, 33.75] | 0.551 |
| HAPO | | | | 0.293 |
| No | 233 (78%) | 127 (80%) | 106 (75%) | |
| Yes | 67 (22%) | 31 (20%) | 36 (25%) | |
| Week at delivery | 39 [38, 40] | 39 [38, 40] | 38 [37, 39] | < 0.001 |
| Admission SBP, mmHg | 126 [116, 139] | 118 [110, 122] | 140 [130, 150] | < 0.001 |
| Admission DBP, mmHg | 76 [69.25, 86] | 70 [65, 75] | 86.5 [79.25, 93] | < 0.001 |
| Prepregnancy BMI, $kg/m^2$ | 21.6[19.78, 23.83] | 20.76[19.19, 22.62] | 22.59 [20.45, 25.89] | < 0.001 |
| BW, g | 3180 [2860, 3470] | 3305[3062.5, 3500] | 2980 [2460, 3410] | < 0.001 |
| Newborn Sex | | | | 0.149 |
| Female | 140 (47%) | 67 (42%) | 73 (51%) | |
| Male | 160 (53%) | 91 (58%) | 69 (49%) | |
| Fetal Outcome | | | | < 0.001 |
| Adverse Outcome | 50 (17%) | 0 (0%) | 50 (35%) | |
| Normal Outcome | 250 (83%) | 158 (100%) | 92 (65%) | |
| LOS, days | 4 [3, 6] | 3[2, 3] | 6[4, 8] | < 0.001 |
| 24h UTP, mg/d | 669.21[407.5, | 114.55[102.28, | 708.91[440.39, | < 0.001 |
| MAU, mg/L | 350 [172.5, 795] | 59.78[51.63, 70.31] | 367.19 [190, 815] | < 0.001 |
| Perinatal Anemia | | | | < 0.001 |
| Normal | 179 (60%) | 108 (68%) | 71 (50%) | |
| Mild | 100 (33%) | 46 (29%) | 54 (38%) | |
| Moderate | 21 (7%) | 4 (3%) | 17 (12%) | |

Continuous variables are reported as median [interquartile range], and categorical variables as n (%). Descriptive delivery and neonatal outcome variables are shown for cohort characterization and were not used as prediction features.
Abbreviations: P-TMA, pregnancy-associated thrombotic microangiopathy; HAPO, history of adverse pregnancy outcomes; SBP, systolic blood pressure; DBP, diastolic blood pressure; BMI, body mass index; BW, birth weight; LOS, length of hospital stay; 24h UTP, 24-hour urinary total protein; MAU, microalbuminuria.

Stratified sampling assigned 240 participants to the training cohort and 60 participants to the held-out test cohort. The P-TMA proportion was similar in the training cohort (47.5%) and held-out test cohort (46.7%), supporting balanced internal evaluation. After preprocessing, 146 longitudinal laboratory predictors were retained for model development.

### 3.2 Model performance in the held-out test cohort

Gradient boosting was selected as the final model according to the prespecified training-cohort cross-validation rule. In the held-out test cohort, the selected model achieved an AUROC of 0.872 and an AUPRC of 0.883. At the threshold derived from out-of-fold training predictions, the model achieved accuracy of 0.783, sensitivity of 0.750, specificity of 0.812, PPV of 0.778, NPV of 0.788, and F1 score of 0.764 (Table 2). Bootstrap 95% confidence intervals were 0.769–0.952 for AUROC and 0.780–0.959 for AUPRC. The Brier score was 0.185.

**Table 2:** Held-out test performance of the selected gradient boosting model.

| Metric | Estimate (bootstrap 95% CI) |
|---|---|
| AUROC | 0.872 (0.769–0.952) |
| AUPRC | 0.883 (0.780–0.959) |
| Accuracy | 0.783 (0.683–0.883) |
| Sensitivity | 0.750 (0.583–0.897) |
| Specificity | 0.812 (0.676–0.938) |
| PPV | 0.778 (0.600–0.931) |
| NPV | 0.788 (0.643–0.914) |
| F1 score | 0.764 (0.615–0.877) |
| Brier score | 0.185 (0.100–0.279) |
| ECE | 0.193 (0.103–0.293) |

Confidence intervals were estimated using 1000 bootstrap resamples of the held-out test cohort. AUPRC, area under the precision-recall curve; AUROC, area under the receiver operating characteristic curve; ECE, expected calibration error; NPV, negative predictive value; PPV, positive predictive value. Brier score and ECE are lower-is-better metrics.

Tree-based ensemble models showed the strongest overall held-out test performance. Extra trees had the highest numerical test AUROC, but gradient boosting remained the selected reporting model because it had been chosen a priori within the training workflow (Figure 3). This preserves the held-out test cohort as the final evaluation set and avoids post hoc test-set-driven model selection.

Figure 2 summarizes test-cohort discrimination, precision-recall performance, calibration, and threshold-based classification. The row-normalized confusion matrix showed correct classification of 75.0% of observed P-TMA cases and 81.2% of observed controls, indicating balanced recognition of cases and controls rather than performance driven by one outcome class alone.

**Table 3:** Held-out test performance of candidate machine learning models.

| Model | AUROC | AUPRC | Accuracy | Sensitivity | Specificity | PPV | NPV | F1 score | Brier score |
|---|---|---|---|---|---|---|---|---|---|
| Gradient boosting∗ | 0.872 | 0.883 | 0.783 | 0.750 | 0.812 | 0.778 | 0.788 | 0.764 | 0.185 |
| Extra trees | 0.896 | 0.914 | 0.833 | 0.786 | 0.875 | 0.846 | 0.824 | 0.815 | 0.165 |
| Random forest | 0.863 | 0.891 | 0.783 | 0.679 | 0.875 | 0.826 | 0.757 | 0.745 | 0.190 |
| Logistic regression | 0.834 | 0.833 | 0.733 | 0.571 | 0.875 | 0.800 | 0.700 | 0.667 | 0.172 |
| SVM (RBF) | 0.798 | 0.812 | 0.700 | 0.786 | 0.625 | 0.647 | 0.769 | 0.710 | 0.182 |

∗ Prespecified selected model based on mean five-fold cross-validated AUROC in the training cohort. Candidate-model test results are shown to describe comparative internal performance; the held-out test cohort was not used to revise the selected model. AUPRC, area under the precision-recall curve; AUROC, area under the receiver operating characteristic curve; NPV, negative predictive value; PPV, positive predictive value; SVM, support vector machine. Brier score is lower-is-better.

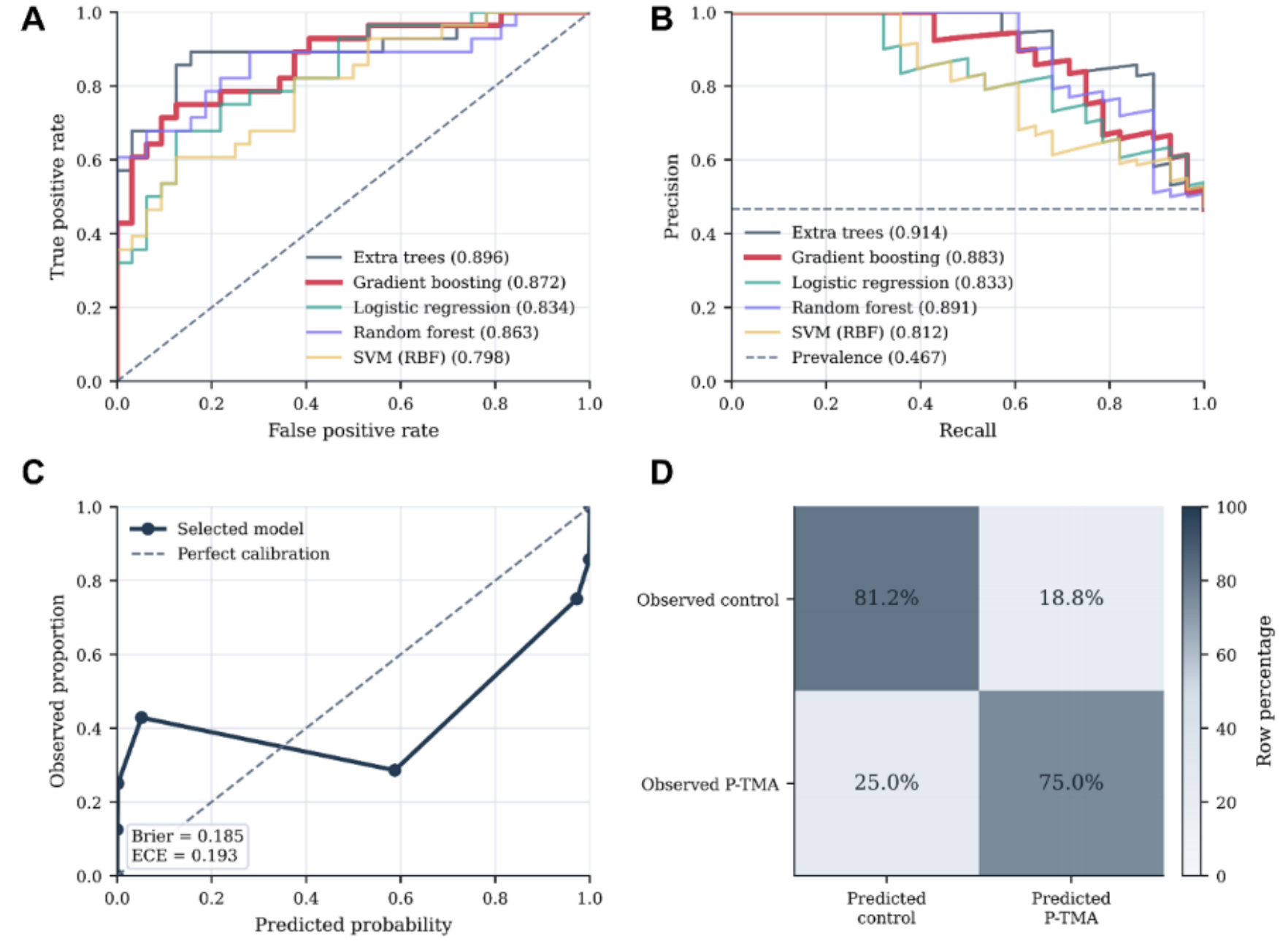

**Figure 2: Held-out test performance of candidate models and the selected model.** (A) ROC curves. (B) Precision-recall curves. (C) Calibration curves. (D) Row-normalized confusion matrix.

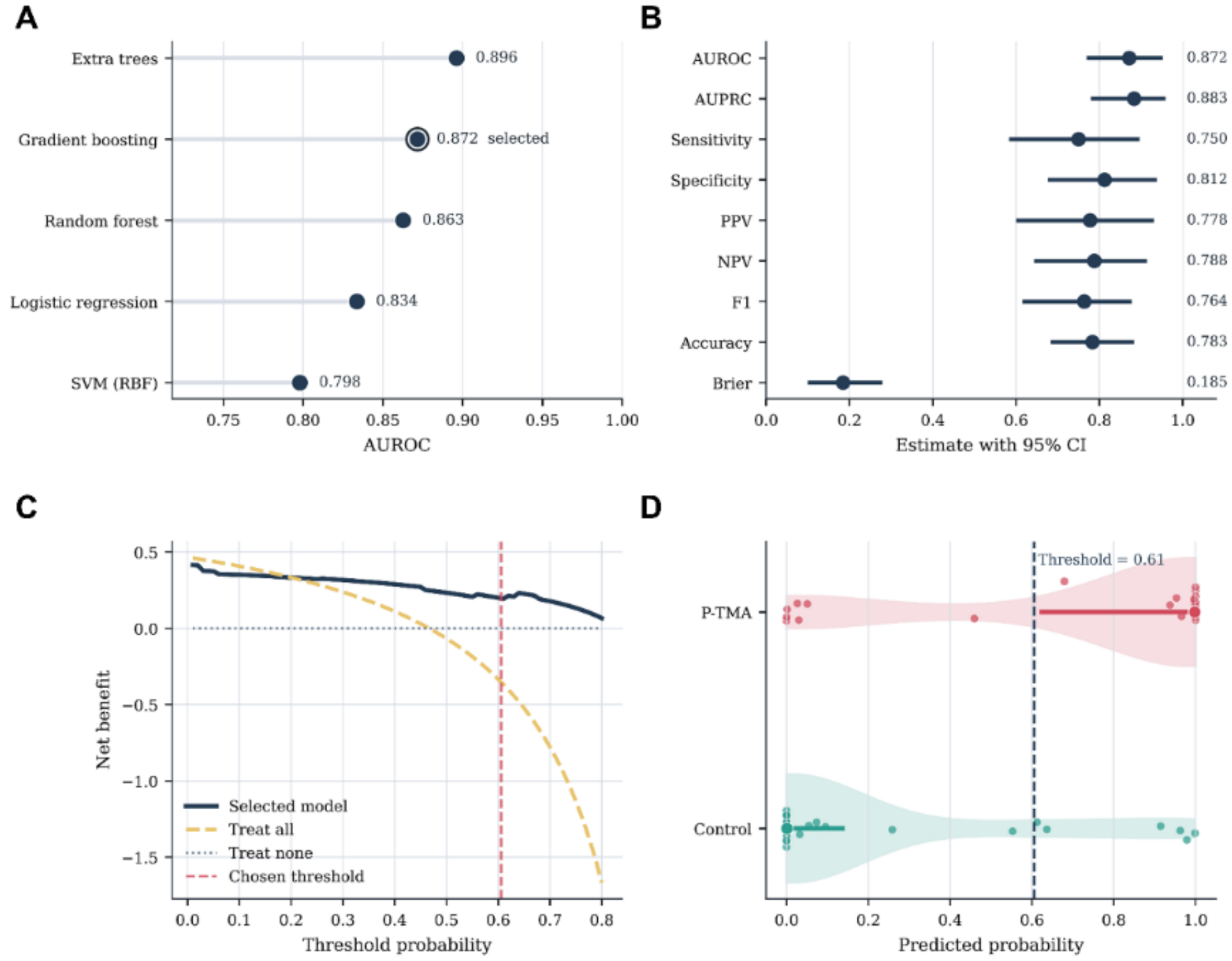

**Figure 3: Held-out test utility and selected-model performance.** (A) Held-out test AUROC across candidate models. (B) Selected model (Gradient boosting). (C) Decision curve analysis. (D) Predicted probability distribution in the held-out test cohort.

### 3.3 Clinical utility and risk

Decision-curve analysis demonstrated positive net benefit for the selected model across a clinically relevant range of threshold probabilities compared with treat-all and treat-none strategies (Figure 3). The predicted-risk distribution was shifted upward among P-TMA cases while retaining partial overlap with controls. This pattern is appropriate for a surveillance-oriented model intended to prioritize closer review rather than to serve as a stand-alone diagnostic rule.

### 3.4 Interpretable laboratory signatures

The selected model identified a coherent set of leading predictors spanning renal-function-related, biochemical injury, inflammatory, urinary, and hematologic domains (Table 4, Figure 4A-B). The top-ranked predictor was CysCVI, corresponding to cystatin C at week 6. Other leading predictors included UAVI (uric acid at week 6), HbXXXII (hemoglobin at week 32), NeuXVI (neutrophils at week 16), LDHVI (lactate dehydrogenase at week 6), and CrXXIV (creatinine at week 24). This profile suggests that the model did not rely on a single laboratory abnormality, but integrated early renal and biochemical signals with later hematologic and inflammatory patterns.

**Table 4:** Leading predictors of the selected gradient boosting model.

| Rank | Feature code | Clinical alias | Importance | Domain | *P* value | Group trend |
|---|---|---|---|---|---|---|
| 1 | CysCVI | Cystatin C at week 6 | 0.157 | Renal/urinary | <0.001 | Upward shift |
| 2 | UAVI | Uric acid at week 6 | 0.119 | Renal/urinary | <0.001 | Higher in P-TMA |
| 3 | HbXXXII | Hemoglobin at week 32 | 0.068 | Hematologic | 0.060 | Lower in P-TMA |
| 4 | NeuXVI | Neutrophils at week 16 | 0.038 | Inflammatory | 0.003 | Higher in P-TMA |
| 5 | LDHVI | Lactate dehydrogenase at week 6 | 0.031 | Biochemical injury | <0.001 | Higher in P-TMA |
| 6 | CrXXIV | Creatinine at week 24 | 0.030 | Renal/urinary | 0.088 | Higher in P-TMA |
| 7 | UWBCXXX | Urine white blood cells at week 30 | 0.030 | Urinary | 0.451 | Limited separation |
| 8 | NeuVI | Neutrophils at week 6 | 0.028 | Inflammatory | <0.001 | Higher in P-TMA |
| 9 | NLRXXIV | Neutrophil-to-lymphocyte ratio at week 24 | 0.028 | Inflammatory | 0.246 | Lower in P-TMA |
| 10 | WBCXII | White blood cell count at week 12 | 0.027 | Hematologic | 0.169 | Higher in P-TMA |

Importance values are model-native feature-importance estimates from the selected gradient boosting model. Feature codes preserve the original longitudinal naming convention; Roman numerals denote gestational timepoints. The group trend summarizes the direction of the distributional pattern shown in Figure 4 and does not imply causality. P-TMA, pregnancy-associated thrombotic microangiopathy.

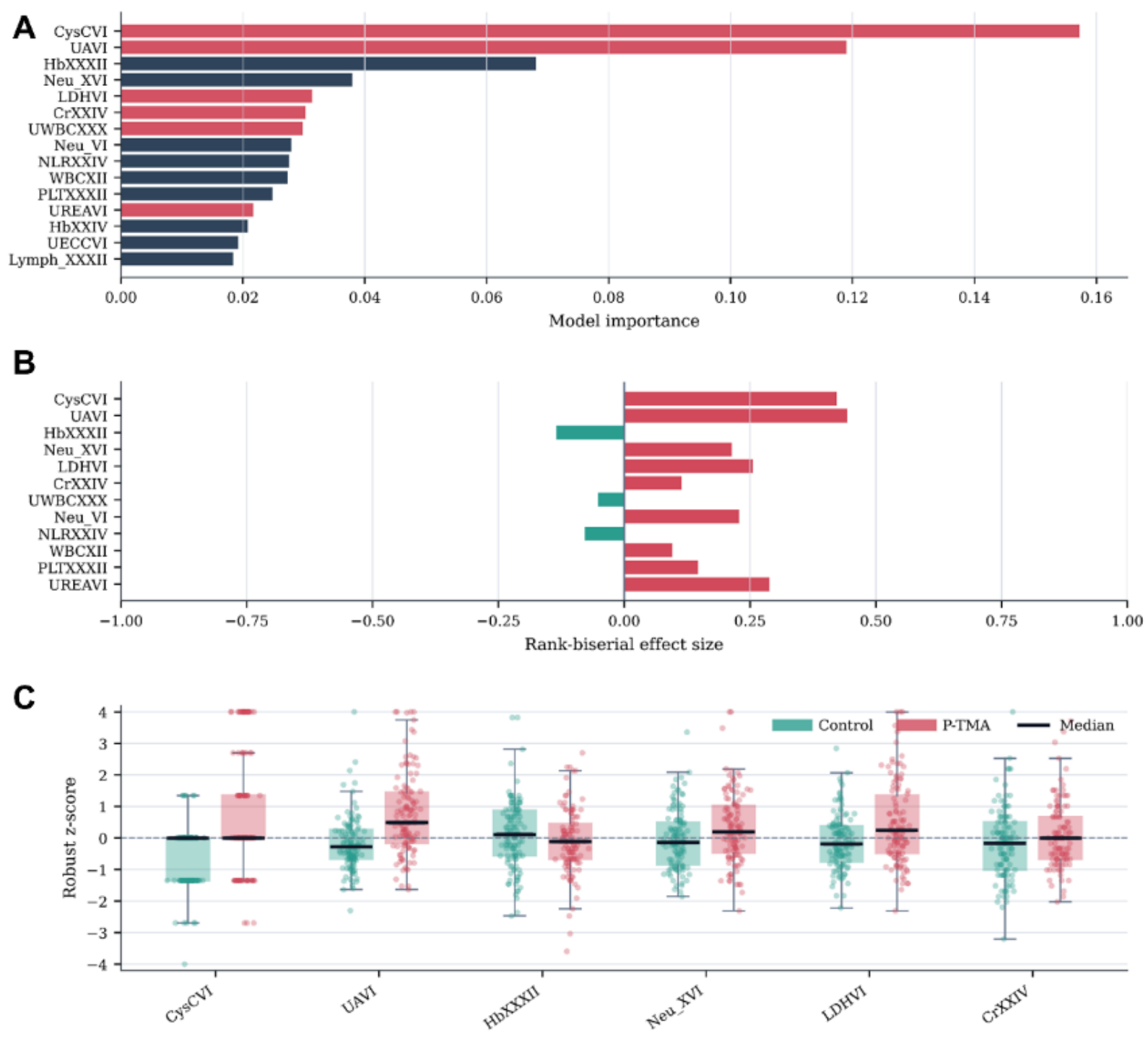

**Figure 4: Model interpretation and leading laboratory signatures.** (A) Top predictors of the selected model. (B) Direction of group difference for leading predictors. (C) Standardized distributions of leading predictors.

Figure 4 describs the direction of group-level distributional differences, and standardized distributions of leading predictors by outcome group. Predictor profiles are shown as robust z-scores to support comparison across laboratory tests with different units and measurement ranges.

## 4 Discussion

Integrating machine learning into disease prediction models based on routine laboratory data has shown great potential[14-16,30]. In this single-center retrospective study, we developed and internally validated an interpretable gradient boosting model that leveraged routine longitudinal laboratory data for risk of P-TMA. Compared with existing prediction models, our approach demonstrated superior performance in P-TMA risk prediction. Farajollahi et al. built diagnostic models for HELLP syndrome using data mining, achieving F1 scores >99% with multilayer perceptron and deep learning[26]; however, those models were based on cross-sectional data from symptomatic patients, not designed for longitudinal risk screening in asymptomatic antepartum women. Tang et al. validated the PLASMIC score for diagnosing TTP in a TMA patient cohort, and combining the score with the LDH/AST ratio raised the positive predictive value to 71% in the pregnant subgroup[31]; yet the PLASMIC score is a rule-based clinical tool that depends on predefined thresholds and cannot capture high-dimensional nonlinear interactions. Li and Wang et al. modeled preeclampsia trajectories using electronic medical record data, achieving antepartum AUROCs of 0.82–0.92[32]; but their study focused on a single disease (preeclampsia) and did not cover the full heterogeneous P-TMA spectrum. In our study, the selected model showed favorable discrimination in a held-out test set (AUROC 0.872, AUPRC 0.883, Brier score 0.185), with balanced sensitivity (0.750) and specificity (0.812) and robust positive and negative predictive values (PPV 0.778, NPV 0.788), while calibration requires further evaluation and potential recalibration in external cohorts. More importantly, our model is the first gradient boosting model to integrate multi-time-point routine laboratory data from gestational weeks 6 to 32, targeted asymptomatic antepartum women, and cover the full P-TMA spectrum, thereby filling a critical gap in the literature. Here we developed and internally validated an interpretable gradient boosting model leveraging routine longitudinal laboratory data for early P-TMA risk (Cystatin C at gestational week 6). To the best of our knowledge, this work provides the first systematic evidence that predictive information for P-TMA was extractable from routine antenatal laboratory profiles.

Early P-TMA eludes detection because its incipient signals are distributed across multiple physiological axes rather than concentrated in a single analyte. Our model captured this multidimensional architecture, integrating early renal function markers, biochemical injury indices, inflammatory features, urinary variables, and later hematologic changes—a profile mechanistically consonant with the endothelial injury, microvascular thrombosis, and renal-hematologic disturbances that define P-TMA[1,2]. The resulting predictor landscape traces a pathophysiological trajectory: first-trimester renal signals (cystatin C and uric acid at week 6), second-trimester inflammatory shifts (neutrophils at week 16 and NLR at week 24), gestational biochemical injury markers (lactate dehydrogenase), and renal function alterations with anemia (creatinine at week 24, hemoglobin at week 32). This “renal–inflammatory–hematologic” cascade aligns with the complement alternative pathway dysregulation that drives endothelial damage, platelet activation, and microthrombosis[3,10].

Among all candidate predictors, cystatin C at week 6 of gestation (CysCVI) ranked as the strongest individual contributor to P-TMA risk. Cystatin C detects subtle glomerular filtration impairment earlier than creatinine[33,34], and its importance in the model arose from distributional shifts and interactions with other laboratory variables rather than from a single cutoff. This positions CysCVI as an integrative early renal risk signal within a multivariable framework, not a stand-alone biomarker. The observation supports the hypothesis that subclinical glomerular hemodynamic or filtration changes—possibly linked to endothelial dysfunction—may already be present in the first trimester among women destined to develop P-TMA, alterations imperceptible through routine assessment[35]. Cystatin C may therefore offer a path toward early risk surveillance.

Methodologically, we adhered to a conservative model selection rule. Although extra trees numerically topped the held-out AUROC, gradient boosting was chosen a priori by cross-validation within the training cohort, preserving the integrity of the test set and conforming to TRIPOD guidelines[36,37]. The comparable performance across tree-based ensembles indicates that predictive signal resides in the data architecture, not in a particular algorithm—a finding consistent with the principle that data define the ceiling of achievable performance and that leaves room for future model refinement with architectures like TabNet or AutoGluon[38,39].

This framework envisions a new decision-support paradigm. Definitive diagnostics—ADAMTS13 activity, complement genetics, renal biopsy—are too slow or costly for universal deployment. Yet routine antenatal phlebotomy is ubiquitous. An embedded model within the electronic health record could continuously generate risk scores without incremental cost or clinician effort[40]. When risk exceeds a threshold, the system could prompt intensified blood pressure and proteinuria surveillance, repeat renal and hematologic testing, or multidisciplinary consultation. Such model-driven active surveillance could compress the 'cognition-to-action' interval that separates nonspecific presentation from specialist intervention—a critical window in atypical hemolytic uremic syndrome, where initiation of C5 blockade (e.g., eculizumab) within 48 hours can rescue renal function[41-43].

The translational advantages are evident: all inputs are laboratory tests already obtained in routine obstetric care, requiring no imaging, omics, or experimental assays. Model outputs are interpretable in terms of familiar variables—cystatin C, uric acid, lactate dehydrogenase, creatinine, hemoglobin, neutrophils—lowering the barrier for clinical review and integration into electronic dashboards or risk-monitoring workflows.

This study is inherently limited by its single-center retrospective design; calibration drift remains a risk when the model encounters different institutions, populations, or assay platforms[44,45]. The sample, although substantial for a rare disease (142 cases), mandates caution for extremely low-incidence subtypes like TTP, and prospective validation with subgroup analyses is necessary[46]. Moreover, the fixed-time-window 'snapshot' approach discards temporal dynamics; future models should exploit architectures such as LSTMs or Transformers that natively handle irregular sampling and dynamic trajectories[47].

Priorities for translational development include: (1) rigorous multicenter, prospective external validation—potentially through global TMA registries—augmented by privacy-preserving federated learning to refine the model without exposing patient-level data[48]; (2) construction of dynamic, personalized baseline-deviation models that set individualized risk thresholds from early-pregnancy laboratory values to mitigate false positives[49]; and (3) cluster randomized trials comparing model-guided surveillance against usual care on definitive endpoints (time to complement inhibitor, ICU admission, maternal-fetal mortality) with accompanying health economic evaluation.

In conclusion, we present a transparent, clinically grounded machine-learning framework that transforms serial routine laboratory data into actionable risk for P-TMA. The interpretable model not only demonstrates robust internal performance but also reveals a predictor signature coherent with disease pathogenesis, establishing a methodological foundation for earlier identification of high-risk pregnancies and for informing perinatal surveillance strategies.

## 5 Conclusion

This study demonstrates that interpretable machine learning can extract clinically meaningful risk signals for P-TMA from serial laboratory tests obtained during routine care. The selected gradient boosting model achieved strong held-out test performance and identified a coherent predictor profile spanning early renal-function-related markers, biochemical injury signals, inflammatory features, and later hematologic abnormalities. These results suggest that longitudinal clinical laboratory patterns may support earlier recognition of patients at increased P-TMA risk and help prioritize closer surveillance before overt clinical deterioration. Further prospective evaluation will be important to confirm clinical utility and guide integration into obstetric decision-support workflows.

## Ethics approval

This study was conducted in accordance with the ethical principles outlined in the Declaration of Helsinki. Written informed consent was obtained from the participants.

## Acknowledgments

**Author contributions:** C.S. and Z. Y. were responsible for conceptualization and study design, methodology development, formal analysis, writing the original draft, and visualization. C.S. and Q. F. performed data curation, and participated in writing, review and editing. Q. C. and F. Y. conducted the investigation, provided essential resources, supervised the project, handled project administration, and acquired funding. All authors have read and approved the final version of the manuscript.

**Funding:** This work was supported by Peking University International Hospital Research Grant (No.YN2020QN01, No.YN2022QX01, and No.YN2020ZD03, No.YN2023ZD01), the Capital's Funds for Health Improvement and Research (No.2024-2-8021) and the Beijing High-Level Innovation and Entrepreneurship Talent Support Program Leading Talent Projects (No. G202511080).

**Competing interests:** The authors declare that there is no conflict of interest regarding the publication of this article.

## Data availability

The data that support the findings of this study are available from the corresponding author upon reasonable request.

## References

[1] Urra M, Lyons S, Teodosiu C G, et al. Thrombotic microangiopathy in pregnancy: Current understanding and management strategies[J]. Kidney International Reports, 2024, 9(8): 2353-2371.
[2] Fakhouri F, Scully M, Provôt F, et al. Management of thrombotic microangiopathy in pregnancy and postpartum: Report from an international working group[J]. Blood, 2020, 136(19): 2103-2117.
[3] Scully M, Neave L. Etiology and outcomes: Thrombotic microangiopathies in pregnancy[J]. Research and Practice in Thrombosis and Haemostasis, 2023, 7(2): 100084.
[4] George J N, Nester C M. Syndromes of thrombotic microangiopathy[J]. New England Journal of Medicine, 2014, 371(7): 654-666.
[5] Tayyab A, Al-sadi A, Irfan S, et al. Pregnancy-associated atypical hemolytic uremic syndrome: A systematic review and meta-analysis of clinical features, treatment strategies, and maternal–fetal outcomes[J]. Blood, 2025, 146(Supplement 1): 4878.
[6] Bussel J B, Hou M, Cines D B. Management of primary immune thrombocytopenia in pregnancy[J]. New England Journal of Medicine, 2023, 389(6): 540-548.
[7] Ferrari B, Peyvandi F. How I treat thrombotic thrombocytopenic purpura in pregnancy[J]. Blood, 2020, 136(19): 2125-2132.
[8] Brown M A, Magee L A, Kenny L C, et al. Hypertensive disorders of pregnancy: ISSHP classification, diagnosis, and management recommendations for international practice[J]. Hypertension, 2018, 72(1): 24-43.
[9] George J N, Nester C M, McIntosh J J. Syndromes of thrombotic microangiopathy associated with pregnancy[J]. Hematology, 2015, 2015(1): 644-648.
[10] Fakhouri F, Frémeaux-Bacchi V. Thrombotic microangiopathy in aHUS and beyond: Clinical clues from complement genetics[J]. Nature Reviews. Nephrology, 2021, 17(8): 543-553.
[11] Zheng X L, Al-Housni Z, Cataland S R, et al. 2025 focused update of the 2020 ISTH guidelines for management of thrombotic thrombocytopenic purpura[J]. Journal of thrombosis and haemostasis: JTH, 2025, 23(11): 3711-3732.
[12] Sánchez-Luceros A, Farías C E, Amaral M M, et al. von willebrand factor-cleaving protease (ADAMTS13) activity in normal non-pregnant women, pregnant and post-delivery women[J]. Thrombosis and Haemostasis, 2004, 92(6): 1320-1326.
[13] Cines D B, Levine L D. Thrombocytopenia in pregnancy[J]. Hematology: the American Society of Hematology Education Program, 2017, 2017(1): 144-151.
[14] Beam A L, Drazen J M, Kohane I S, et al. Artificial intelligence in medicine[J]. New England Journal of Medicine, 2023, 388(13): 1220-1221.
[15] Wang Z, Lee J W, Chakraborty T, et al. Survival modeling using deep learning, machine learning, and statistical methods: A comparative analysis for predicting mortality after hospital admission[J]. Health Data Science, 2026, 6: 0449.
[16] Sun C, Song M, Cai D, et al. A review of deep learning methods for irregularly sampled medical time series data[J]. Health Data Science, 6: 0456.
[17] Watanabe M, Eguchi A, Sakurai K, et al. Prediction of gestational diabetes mellitus using machine learning from birth cohort data of the japan environment and children's study[J]. Scientific Reports, 2023, 13(1): 17419.

[18] Ranjbar A, Montazeri F, Ghamsari S R, et al. Machine learning models for predicting preeclampsia: A systematic review[J]. BMC Pregnancy and Childbirth, 2024, 24: 6.
[19] Mustafa H J, Kalafat E, Prasad S, et al. Prediction of hypertension and diabetes in twin pregnancy using machine learning model based on characteristics at first prenatal visit: National registry study[J]. Ultrasound in Obstetrics & Gynecology, 2025, 65(5): 613-623.
[20] Devoe L D, Muhanna M, Maher J, et al. Current state of artificial intelligence model development in obstetrics[J]. Obstetrics and Gynecology, 2025, 146(2): 233-243.
[21] Du Y, Fang Z, Jiao J, et al. Application of ultrasound-based radiomics technology in fetal-lung-texture analysis in pregnancies complicated by gestational diabetes and/or pre-eclampsia[J]. Ultrasound in Obstetrics & Gynecology, 2021, 57(5): 804-812.
[22] Lundberg S M, Erion G, Chen H, et al. From local explanations to global understanding with explainable AI for trees[J]. Nature Machine Intelligence, 2020, 2(1): 56-67.
[23] Tonekaboni S, Joshi S, McCradden M D, et al. What clinicians want: Contextualizing explainable machine learning for clinical end use[C]//Proceedings of the 4th Machine Learning for Healthcare Conference. PMLR, 2019: 359-380.
[24] Tjoa E, Guan C. A survey on explainable artificial intelligence (XAI): Toward medical XAI[J]. IEEE transactions on neural networks and learning systems, 2021, 32(11): 4793-4813.
[25] Ferraro P M, Lombardi G, Naticchia A, et al. A STARD-compliant prediction model for diagnosing thrombotic microangiopathies[J]. Journal of Nephrology, 2018, 31(3): 405-410.
[26] Farajollahi B, Sayadi M, Langarizadeh M, et al. Presenting a prediction model for HELLP syndrome through data mining[J]. BMC Medical Informatics and Decision Making, 2025, 25: 135.
[27] Bendapudi P K, Hurwitz S, Fry A, et al. Derivation and external validation of the PLASMIC score for rapid assessment of adults with thrombotic microangiopathies: A cohort study[J]. The Lancet Haematology, 2017, 4(4): e157-e164.
[28] Vickers A J, Elkin E B. Decision curve analysis: A novel method for evaluating prediction models[J]. Medical Decision Making: An International Journal of the Society for Medical Decision Making, 2006, 26(6): 565-574.
[29] Collins G S, Reitsma J B, Altman D G, et al. Transparent reporting of a multivariable prediction model for individual prognosis or diagnosis (TRIPOD): The TRIPOD statement[J]. British Journal of Cancer, 2015, 112(2): 251-259.
[30] Zhang M, Zheng Y, Maidaiti X, et al. Integrating machine learning into statistical methods in disease risk prediction modeling: A systematic review[J]. Health Data Science, 4: 0165.
[31] Tang N, Wang X, Li D, et al. Validation of the PLASMIC score, a clinical prediction tool for thrombotic thrombocytopenic purpura diagnosis, in Chinese patients[J]. Thrombosis Research, 2018, 172: 9-13.
[32] Li S, Wang Z, Vieira L A, et al. Improving preeclampsia risk prediction by modeling pregnancy trajectories from routinely collected electronic medical record data[J]. NPJ digital medicine, 2022, 5(1): 68.
[33] Inker L A, Eneanya N D, Coresh J, et al. New creatinine- and cystatin C–based equations to estimate GFR without race[J]. New England Journal of Medicine, 2021, 385(19): 1737-1749.

[34] Shlipak M G, Matsushita K, Ärnlöv J, et al. Cystatin C versus creatinine in determining risk based on kidney function[J]. The New England Journal of Medicine, 2013, 369(10): 932-943.
[35] Bellos I, Fitrou G, Daskalakis G, et al. Serum cystatin-c as predictive factor of preeclampsia: A meta-analysis of 27 observational studies[J]. Pregnancy Hypertension, 2019, 16: 97-104.
[36] Collins G S, Moons K G M, Dhiman P, et al. TRIPOD+AI statement: Updated guidance for reporting clinical prediction models that use regression or machine learning methods[J]. The BMJ, 2024, 385: e078378.
[37] Efthimiou O, Seo M, Chalkou K, et al. Developing clinical prediction models: A step-by-step guide[J]. BMJ, 2024, 386: e078276.
[38] Borisov V, Leemann T, Sebler K, et al. Deep neural networks and tabular data: A survey[J]. IEEE transactions on neural networks and learning systems, 2024, 35(6): 7499-7519.
[39] Topol E J. High-performance medicine: The convergence of human and artificial intelligence[J]. Nature Medicine, 2019, 25(1): 44-56.
[40] Obermeyer Z, Emanuel E J. Predicting the future - big data, machine learning, and clinical medicine[J]. The New England Journal of Medicine, 2016, 375(13): 1216-1219.
[41] Legendre C M, Licht C, Muus P, et al. Terminal complement inhibitor eculizumab in atypical hemolytic-uremic syndrome[J]. The New England Journal of Medicine, 2013, 368(23): 2169-2181.
[42] Morales E, Galindo A, García L, et al. Eculizumab in early-stage pregnancy[J]. Kidney International Reports, 2020, 5(12): 2383-2387.
[43] Che M, Moran S M, Smith R J, et al. A case-based narrative review of pregnancy-associated atypical hemolytic uremic syndrome/complement-mediated thrombotic microangiopathy[J]. Kidney international, 2024, 105(5): 960-970.
[44] Wiens J, Saria S, Sendak M, et al. Do no harm: A roadmap for responsible machine learning for health care[J]. Nature Medicine, 2019, 25(9): 1337-1340.
[45] Zech J R, Badgeley M A, Liu M, et al. Variable generalization performance of a deep learning model to detect pneumonia in chest radiographs: A cross-sectional study[J]. PLoS Medicine, 2018, 15(11): e1002683.
[46] Scully M, Yarranton H, Liesner R, et al. Regional UK TTP registry: Correlation with laboratory ADAMTS 13 analysis and clinical features[J]. British Journal of Haematology, 2008, 142(5): 819-826.
[47] Keshari R, Vatsa M, Singh R, et al. Learning structure and strength of CNN filters for small sample size training[C]//2018 IEEE/CVF Conference on Computer Vision and Pattern Recognition. 2018: 9349-9358.
[48] chander B, John C, Warrier L, et al. Toward trustworthy artificial intelligence (TAI) in the context of explainability and robustness[J]. ACM Comput. Surv., 2025, 57(6): 144:1-144:49.
[49] Ghassemi M, Naumann T, Schulam P, et al. A review of challenges and opportunities in machine learning for health[J]. AMIA Summits on Translational Science Proceedings, 2020, 2020: 191-200.